%% file: iccv.tex
\documentclass[10pt,twocolumn,letterpaper]{article}

\usepackage{iccv}
\usepackage{times}
\usepackage{epsfig}
\usepackage{graphicx}
\usepackage{amsmath}
\usepackage{amssymb}
\usepackage{marvosym}
\usepackage{adjustbox}
\usepackage{overpic}
\usepackage{soul}
\usepackage{multirow, colortbl, xcolor, booktabs}
\usepackage{rotating}
\usepackage{bm}
\usepackage{algorithm}
\usepackage{listings}
\usepackage{algorithmic}  
\usepackage{newfloat}
\usepackage[accsupp]{axessibility}
\usepackage[linesnumbered, algo2e]{algorithm2e}
\usepackage{xcolor} 
\newcommand{\nn}{USAGE}
\definecolor{Highlight}{HTML}{39b54a}
\definecolor{airforceblue}{rgb}{0.36, 0.54, 0.66}

\definecolor{Blue3}{RGB}{96, 158, 214}
\definecolor{Blue1}{RGB}{214, 235, 245}
\definecolor{Blue2}{RGB}{235, 245, 250}
\definecolor{Red1}{RGB}{244,178,133}
\usepackage[pagebackref=true,breaklinks=true,colorlinks,bookmarks=false]{hyperref}

\iccvfinalcopy % *** Uncomment this line for the final submission

\def\iccvPaperID{1340} % *** Enter the ICCV Paper ID here

\ificcvfinal\fi
\input{sec/preamble}
\begin{document}

%%%%%%%%% TITLE
\title{USAGE: A Unified Seed Area Generation Paradigm for Weakly Supervised Semantic Segmentation}

\author{Zelin Peng$^{1}$, Guanchun Wang$^{2}$, Lingxi Xie$^{3}$, Dongsheng Jiang$^{3}$, Wei Shen$^{1{(\textrm{\Letter})}}$, and Qi Tian$^{3}$\\
	$^1$MoE Key Lab of Artificial Intelligence, AI Institute, Shanghai Jiao Tong University\\ $^2$School of Artificial Intelligence, Xidian University $^3$Huawei Inc.\\
	{\tt\small 
		\{zelin.peng,  wei.shen\}@sjtu.edu.cn;  gwang\underline{~}2@stu.xidian.edu.cn;} \\ {\tt\small 198808xc@gmail.com; dongsheng\underline{~}jiang@outlook.com; tian.qi1@huawei.com}
}

\maketitle
\ificcvfinal\thispagestyle{empty}\fi

\let\thefootnote\relax\footnote{$^{\textrm{\Letter}}$Corresponding Author.}

\input{sec/0_abstract}
\input{sec/1_introduction}

% \clearpage
\input{sec/2_related}

\input{sec/3_method}

\input{sec/4_results}

\input{sec/5_conclusions}

{\small
\bibliographystyle{ieee_fullname}
\bibliography{iccv}
}

\end{document}

%% file: sec/preamble.tex
\usepackage{overpic}
\usepackage{enumitem} %< control spacing in itemize/enumerate/...
\usepackage{overpic} %< add raw math symbols to figures
\usepackage{color}
\usepackage{amssymb}
\usepackage{pifont}
\usepackage{bbding}
\newcommand{\cmark}{\ding{51}}
\definecolor{turquoise}{cmyk}{0.65,0,0.1,0.3}
\definecolor{purple}{rgb}{0.65,0,0.65}
\definecolor{dark_green}{rgb}{0, 0.5, 0}
\definecolor{orange}{rgb}{0.8, 0.6, 0.2}
\definecolor{red}{rgb}{0.8, 0.2, 0.2}
\definecolor{darkred}{rgb}{0.6, 0.1, 0.05}
\definecolor{blueish}{rgb}{0.0, 0.3, .6}
\definecolor{light_gray}{rgb}{0.7, 0.7, .7}
\definecolor{pink}{rgb}{1, 0, 1}
\definecolor{greyblue}{rgb}{0.25, 0.25, 1}

\usepackage{blindtext}

\renewcommand{\paragraph}[1]{\vspace{1em}\noindent\textbf{#1}.}

%% file: sec/0_abstract.tex
\begin{abstract}

Seed area generation is usually the starting point of weakly supervised semantic segmentation (WSSS). Computing the Class Activation Map (CAM) from a multi-label classification network is the de facto paradigm for seed area generation, but CAMs generated from Convolutional Neural Networks (CNNs) and Transformers are prone to be under- and over-activated, respectively, which makes the strategies to refine CAMs for CNNs usually inappropriate for Transformers, and vice versa. In this paper, we propose a \textbf{U}nified optimization paradigm for \textbf{S}eed \textbf{A}rea \textbf{GE}neration (USAGE) for both types of networks, in which the objective function to be optimized consists of two terms: One is a generation loss, which controls the shape of seed areas by a temperature parameter following a deterministic principle for different types of networks; The other is a regularization loss, which ensures the consistency between the seed areas that are  generated by self-adaptive network adjustment from different views, to overturn false activation in seed areas. Experimental results show that USAGE consistently improves seed area generation for both CNNs and Transformers by large margins, e.g., outperforming state-of-the-art methods by a mIoU of 4.1\% on PASCAL VOC. Moreover, based on the USAGE-generated seed areas on Transformers, we achieve state-of-the-art WSSS results on both PASCAL VOC and MS COCO.

\end{abstract}

%% file: sec/1_introduction.tex
%\vspace{-5pt}
\section{Introduction}
\label{sec:intro}

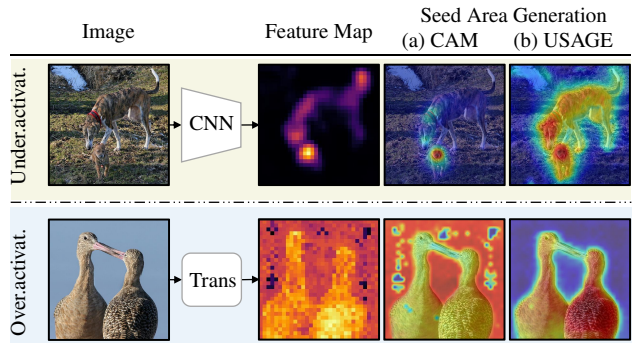
\begin{figure}[t]
	\centering
	%\fbox{\rule{0pt}{0.5in} \rule{0.9\linewidth}{0pt}}
	\footnotesize
	\begin{overpic}[width=0.99\linewidth]{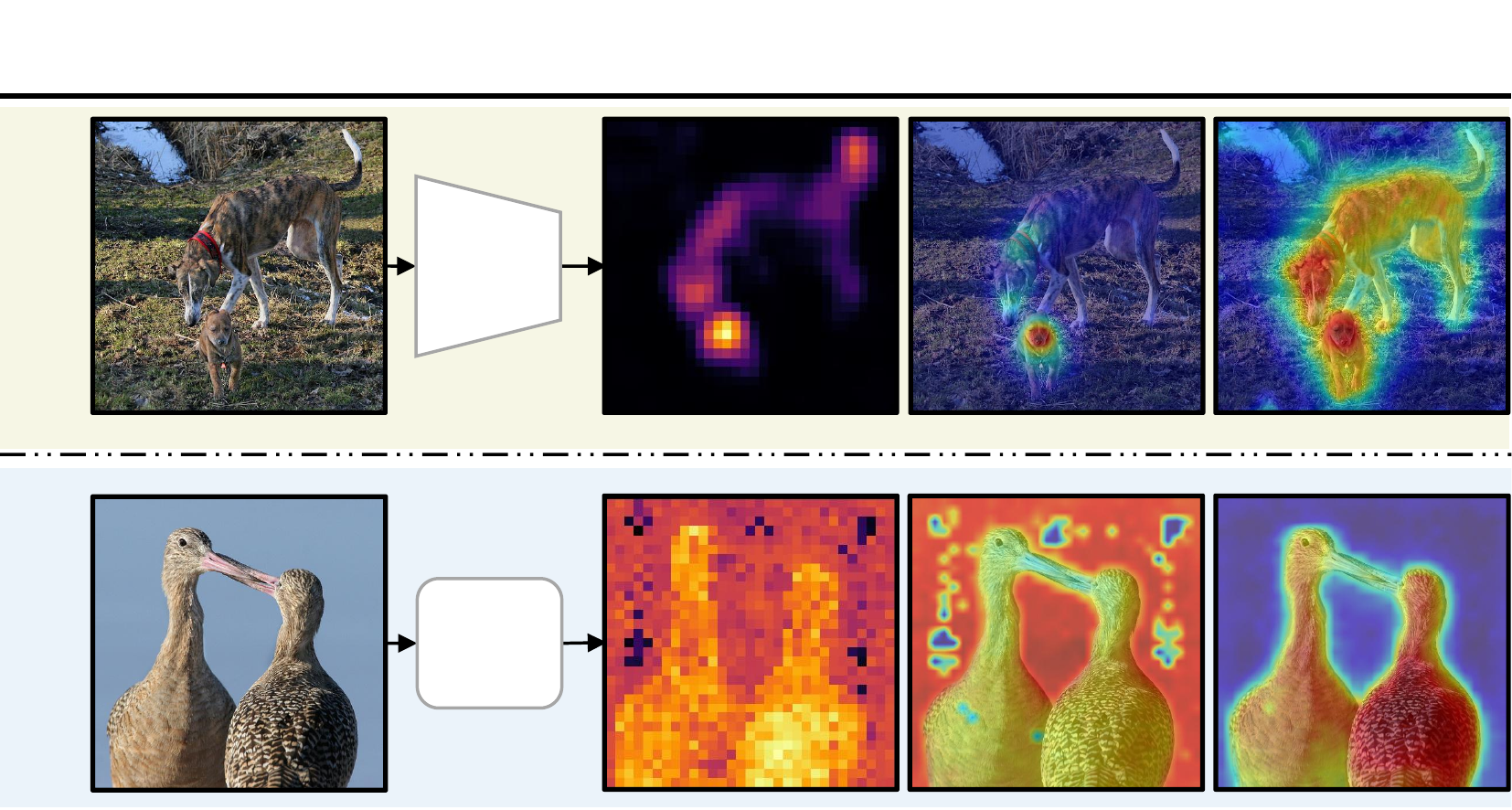}

        \put(28.8,35.1){CNN}
	\put(28.8,9.8){Trans}
		\put(11.5,50){Image}
		\put(41,50){Feature Map}
		\put(63,48.5){(a) CAM}
		\put(66,52.5){Seed Area Generation}
		\put(81,48.5){(b) USAGE}

        \rotatebox{90}{\put(25.5,12){ Under.activat.}{\put(2,12){Over.activat.}}}
	\end{overpic}

       \caption{Seed area visualization. (a) CAM-based seed areas generated from a CNN (ResNet38~\cite{SS_ResNet38_2019_PR}) and a Transformer (DeiT-S~\cite{Trans_tde_2021_ICML}). (b) USAGE-based seed areas (ours) generated from the CNN and the Transformer.}

	\label{fig:fig1}
\end{figure}

\par 

The goal of weakly supervised semantic segmentation (WSSS) is to train a semantic segmentation model under weak supervision, \emph{i.e.}, image-level labels, so that the burden of relying on pixel-level labels is largely reduced. Among various types of weak supervision, we focus on studying WSSS under image-level labels~\cite{WSSS_ICD_2020_CVPR,WSSS_SEC_2016_ECCV,WSSS_DSRG_2018_CVPR,WSSS_OAA_2019_ICCV,WSSS_PPC_2022_CVPR, WSSS_HAS_2017_ICCV}, which is considered one of the most challenging scenarios and has attracted increasing attention. 

Seed area generation is usually the first step of WSSS, which produces an initial pseudo mask based on the image-level labels on each training image. This is often achieved by first optimizing a mapping between the dense feature map of the image and the image-level labels (\emph{i.e.,} training a deep neural network for multi-class classification), and then inferring the contribution of each location on the feature map to the classification result \emph{w.r.t.} each specific class.

The Class Activation Map (CAM)~\cite{CAM_cam_2018_CVPR} has been the \emph{de facto} paradigm for seed area generation. However, it is known that, based on the CAM, seed areas generated from Convolutional Neural Networks (CNNs) are prone to be under-activated~\cite{CAM_rethink_2020_ECCV}. In addition, when the backbone is changed from CNNs to Transformers (which have been adopted in WSSS very recently and report state-of-the-art performance~\cite{WSSS_MCT_2022_CVPR}), the CAM is inversely prone to result in over-activated seed areas~\cite{WSSS_max_2022_ECCV}, as shown in Fig.~\ref{fig:fig1}. Consequently, it is difficult to design a seed area generation paradigm that is appropriate for both CNN-based and Transformer-based WSSS. For example, the widely used CAM refinement strategy-``seed and expand''~\cite{WSSS_SEC_2016_ECCV} has achieved success in CNN-based WSSS, but it is easy to deteriorate the over-activation of seed areas in Transformer-based WSSS.

In this paper, we propose a \textbf{U}nified optimization paradigm for \textbf{S}eed \textbf{A}rea \textbf{GE}neration (USAGE) for both types of networks, in which the objective function to be optimized consists of two terms: a generation loss and a regularization loss. The generation loss measures the fitness between the contribution of a seed area and the classification result and controls the shape of the seed area by introducing a temperature parameter based on a deterministic principle. This means how to set the temperature parameter for different types of networks is deterministic, \emph{i.e.}, set a small/large temperature for Transformers/CNNs to sharpen/smooth the seed area.

The regularization loss is designed to ensure the consistency between the seed areas that are generated from different views, which can overturn false activation in seed areas. Previous WSSS methods~\cite{WSSS_SEAM_2020_CVPR,WSSS_L2G_2022_CVPR} mainly adopt geometric transformations to generate different views to instantiate such a regularization loss, which has achieved a great success on CNN-based WSSS, but this view generation strategy might not be appropriate for Transformer-based WSSS, since Transformers are insensitive to various geometric transformations~\cite{robust_2021_NIPS}. Besides, geometric transformations, \emph{e.g.,} multi-crop, may generate views that only contain the background region, which may make the consistency optimization  process ill-posed. To alleviate the issue, we propose a self-adaptive network adjustment strategy to generate different views. In particular, we obtain different views by making adjustments to the architecture of the classification network, where the magnitudes of adjustments are determined by the learning status of the network. This strategy is shown to be effective for both types of networks in rectifying erroneously activated seed areas.

Extensive experiments show that USAGE can significantly improve the quality of the seed areas for both CNNs and Transformers. Furthermore, under the paradigm of USAGE, our instantiation for transformers, \emph{i.e.}, applying sharpening and regularization on the seed areas, leads to new state-of-the-art WSSS results on both the PASCAL VOC~\cite{dataset_pascal_2010_IJCV} and MS COCO~\cite{dataset_coco_2014_ECCV} dataset.

%% file: sec/2_related.tex
\section{Related Work}
\label{sec:related}

WSSS methods can be categorized into two types: step-wise~\cite{WSSS_MCT_2022_CVPR,WSSS_SCE_2020_CVPR,WSSS_CP_2021_ICCV,WSSS_rib_2021_NIPS} and end-to-end~\cite{WSSS_afa_2022_CVPR,WSSS_SSSS_2020_CVPR}. Since step-wise methods generally achieve better performance than end-to-end methods, we focus on the former in this paper.
Most step-wise WSSS methods follow such a sequence of steps: seed area generation, pseudo mask generation, and segmentation model training. Seed area generation is the first step for WSSS which provides initial cues to generate pseudo masks for further segmentation network training. The CAM~\cite{CAM_cam_2018_CVPR} is a widely used technique to generate seed area. However, CAM-based seed areas generated from CNNs and Transformers are prone to be under- and over-activated, respectively.

\textbf{Seed Area Generation from CNNs.} 
Prior to the widespread usage of CAMs, earlier methods~\cite{WSSS_fil_2015_CVPR, WSSS_FC_2014_ICLR} explored using multiple instance learning (MIL) for seed area generation. Pinheiro \emph{et al.}~\cite{WSSS_fil_2015_CVPR} utilized LogSumExp-aggregation~\cite{MIL_logsumexp_2004_cup} to aggregate pixel-level predictions in the output layer, generating image-level scores. Due to the limited quality in seed area generation, MIL-based methods became inactive after the popularization of CAM.  Adversarial erasing \cite{WSSS_ORMAE_2017_CVPR, WSSS_SE_2018_NIPS} is a well-known CAM~\cite{CAM_cam_2018_CVPR} expanding strategy that erases the most discriminative region in a CAM to enforce the classification network to activate on other areas. Wei \emph{et al.}~\cite{WSSS_RCD_2018_CVPR} proposed to enlarge seed areas by an ensemble of the CAMs computed using multiple dilated convolutional blocks with different dilation rates. Lee \emph{et al.}~\cite{WSSS_AdvCAM_2021_CVPR} applied an anti-adversarial manner to perturb images along gradients \emph{w.r.t.} the classification loss. Different from the above methods that mitigate under-activation by expanding seed areas via iteratively assembling, some other methods~\cite{WSSS_SEAM_2020_CVPR, WSSS_CP_2021_ICCV} attempted to solve this problem by maintaining consistency between seed areas from different views, jointly optimized with the classification loss. The most representative method is~\cite{WSSS_SEAM_2020_CVPR}, which generated views by different geometric transformations. However, the view consistency strategy under different geometric transformations might not be appropriate for Transformers, as Transformers are insensitive to various geometric transformations. Unlike these methods, the view consistency strategy in our USAGE is based on network adjustment, which is appropriate to both CNNs and Transformers.

\textbf{Seed Area Generation from Transformers.} 
%The above methods are commonly based on CNNs, while 
Following the research line of Transformers, very recently, Xu \emph{et al.}~\cite{WSSS_MCT_2022_CVPR} proposed a multi-class token transformer to generate the class-to-patch attention map as the seed area. Although the class-to-patch attention map provides an upgraded version of the CAM~\cite{CAM_cam_2018_CVPR} (more concrete details are discussed in Sec.~\ref{sec.4.3}), it also suffers from over-activation. 

Our USAGE can address both the over-activation issue for Transformers and the under-activation issue for CNNs. We also show that, both CAM-based and MIL-based seed area generation methods are special cases of USAGE (Sec.~\ref{sec.4.3}).

%% file: sec/3_method.tex
\section{Revisiting CAM in Seed Area Generation} 
The pipeline of the mainstream WSSS methods consists of three sequential steps: 1) Generating seed areas for training images from a multi-label classification network; 2) Refining seed areas to be pseudo masks via affinity propagation, \emph{e.g.}, PSA~\cite{WSSS_psa_2018_CVPR}; 3) Training a segmentation network based on pseudo masks which can be used to perform segmentation on a test image. We mainly focus on the first step, \emph{i.e.}, seed area generation, since the quality of seed areas can directly influence succeeding steps.

We provide a critical review of CAM~\cite{CAM_cam_2018_CVPR} in seed area generation for WSSS regarding to different types of network backbones, \emph{i.e.}, CNN and Transformer and reveal that how it suffers from problematic activations.

\subsection{Preliminary Background}
\label{sec:3.0}
We begin by introducing the learning paradigm of seed area generation. Formally, let $\mathcal{C}$ be a pre-defined category label set, given an input image $\mathbf{I}$ with its image-level label $\mathbf{y} \in \left\{0,1\right\}^{|\mathcal{C}|}$, existing methods first train a neural network $\mathcal{H}=\mathcal{F} \circ \mathcal{G}$, where a feature extractor $\mathcal{F}$ maps $\mathbf{I}$ to a dense feature map $\mathbf{A}=\mathcal{F}\left(\mathbf{I}\right) \in \mathbb{R}^{W \times H \times D}$, and a classifier $\mathcal{G}$, parameterized with $\mathbf{w}$, maps $\mathbf{A}$ to a score vector over pre-defined categories $\mathbf{s}=\mathcal{G}\left(\mathbf{A},\mathbf{w}\right)\in \mathbb{R}^{|\mathcal{C}|}$. $\mathcal{F}$ can be any type of neural networks, \emph{e.g.,} CNN or Transformer.  %\emph{w.r.t.} class $c$. 
Accordingly, the objective function of seed area generation is formulated as:
\begin{equation}
\begin{aligned}
(\mathbf{\hat{A}}, \mathbf{\hat{w}}) &=\underset{\mathbf{A},\mathbf{w}}{\operatorname{argmin}}~\mathcal{L}\left(\mathbf{y}, \mathbf{s}\right) \\
&=\underset{\mathbf{A},\mathbf{w}}{\operatorname{argmin}}~\frac{1}{|\mathcal{C}|} \sum_c\mathcal{L}_{\text{CE}}^c\left(y^c, \sigma\left(s^c\right)\right),%\ell(Y^c, S^c),%\mathcal{L}\left(S^c, A, w\right)%-Y^c \log \left(\sigma\left(g\left(A,w\right)\right)\right)
\label{baseline_objective}
\end{aligned}
\end{equation}
which defines a cross entropy (CE) loss function for the mapping between features and image-level labels. $s^c$ and $y^c$ denote the score and image-level label of class $c$, respectively, and $\sigma$ is the sigmoid function. During inference, the seed area $\mathbf{M}^c$ for each class $c$ is obtained as:
\begin{equation}
M_{i j}^c= \sum_{d=1}^D \hat{w}_d^c \hat{A}_{i j}^d,
\end{equation}
where $(\hat{A})_{i,j}^d$ is the $d$-th dimension feature at position $(i,j)$, $\hat{w}_{d}^c$ is the weight of the classifier corresponding to class $c$ for feature dimension $d$, and $M_{i j}^c$ is the value of the seed area $\mathbf{M}^c$ at position $(i,j)$.  

\input{tab/comp_cnn_trans}
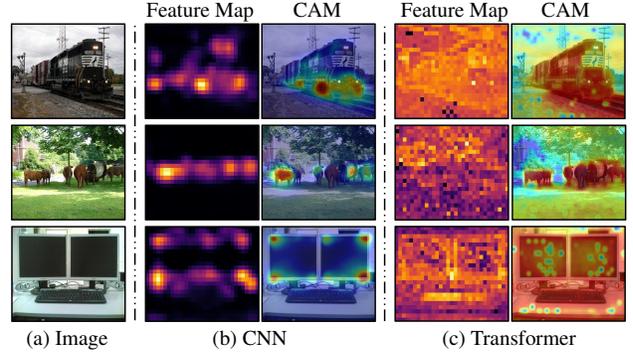
\begin{figure}[t]
	\centering
	%\fbox{\rule{0pt}{0.5in} \rule{0.9\linewidth}{0pt}}
	\footnotesize
	\begin{overpic}[width=0.99\linewidth]{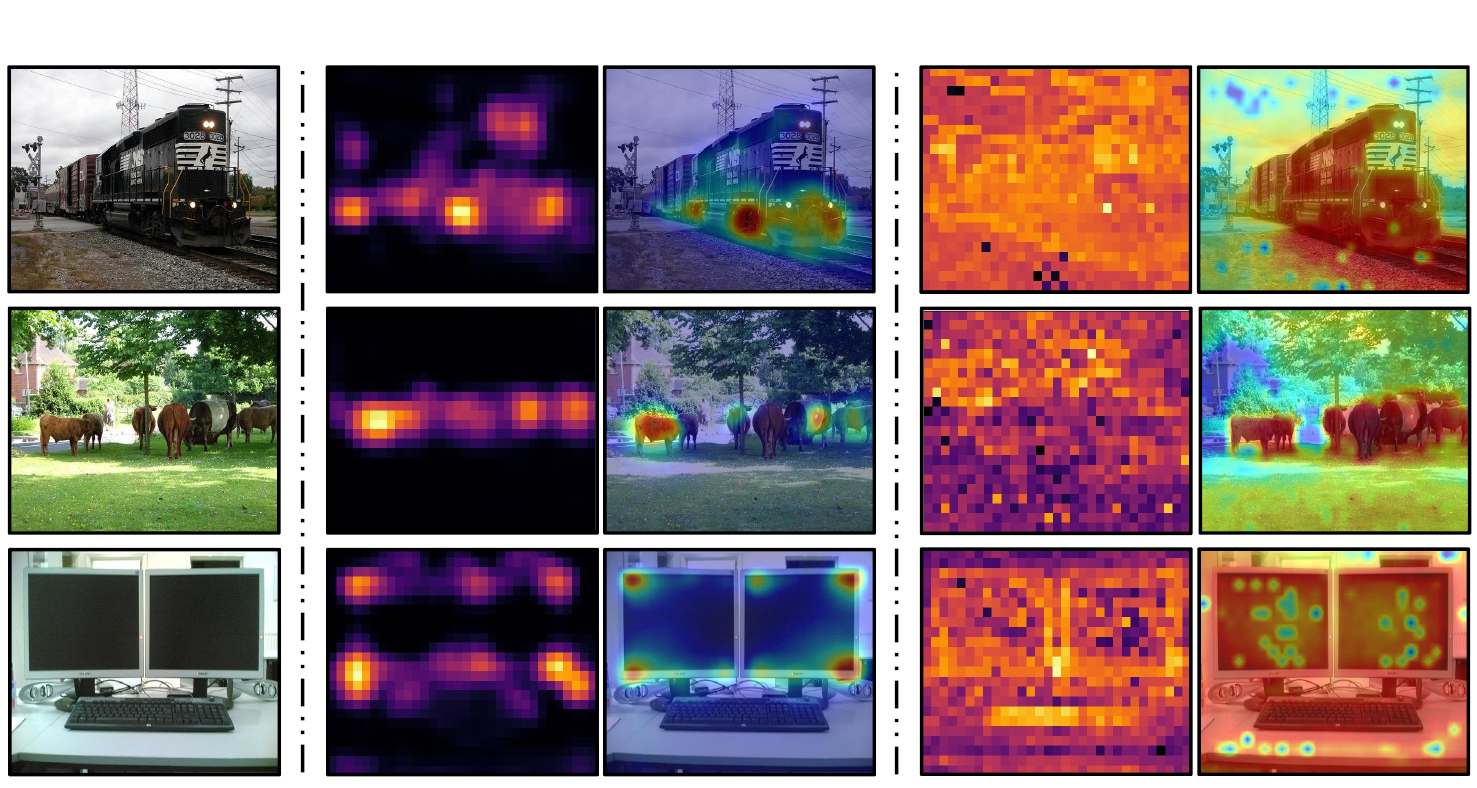}
        \put(3,-3){(a) Image}
        \put(33,-3){(b) CNN}
        \put(70,-3){(c) Transformer}
        \put(21.5,50){ Feature Map}
        \put(46,50){CAM}
        \put(62.5,50){ Feature Map}
        \put(86,50){CAM}
	\end{overpic}
	%\vspace{2mm}
 \vskip 0.1in
	\caption{Comparison of CAMs and feature maps generated from CNNs and Transformers. (a) Input Image. (b) Feature maps and CAMs computed from a CNN. (c) Feature maps and CAMs computed from a Transformer. The above feature maps are averaged along the channel dimension.}
 %\vspace{-4mm}
 \label{cnn_trans_features}
\end{figure}
CAM~\cite{CAM_cam_2018_CVPR} is the \emph{de facto} paradigm for seed area generation, which adopts a global average pooling (GAP) and a fully-connected layer as the classifier. The mapping of the classifier $\mathcal{G}(\cdot,\cdot)$ is specified as:
\begin{equation}
\begin{aligned}
\label{eq:1}
s^c &=\frac{1}{N}\sum_{d=1}^D w_d^c \sum_{i=1}^W \sum_{j=1}^H A_{i j}^d \\
&=\frac{1}{N}\sum_{i=1}^W \sum_{j=1}^H \sum_{d=1}^D w_d^c A_{i j}^d.
%&=\frac{1}{N}\sum_{i=1}^W \sum_{j=1}^H \mathbf{M}_{i j}^c,
\end{aligned}
\end{equation}
where $N$ is the spatial size (= width $W$ × height $H$) of the feature map $\mathbf{A} $.
\subsection{Under-activation and Over-activation}
To determine whether a seed area is over- or under-activated, we first give some metrics derived from the classical definition of false positive and false negative rates. 
\begin{equation}
\text{FPR}=\text{FP}/(\text{T+FP}),\text{FNR}=\text{FN}/(\text{P+FN}),
\end{equation}
where the terms FPR and FNR correspond to false positive rate and false negative rate \emph{w.r.t.} a target class, respectively. T and P refer to true predictions and positive predictions for the target class, respectively. Meanwhile, FP and FN denote false positive predictions and false negative predictions for that class, respectively. 
Intuitively, a higher FNR value implies that more foreground regions are not activated, indicating under-activation. Conversely, a higher FPR value indicates more background regions are mistakenly activated as seed areas, signifying over-activation. Besides, the mean intersection-over-union (mIoU) is used as the typical metric for segmentation.

Table~\ref{tab_cnn_trans} shows quantitative results with three metrics. Compared to the CAM-based seed area generated from the CNN, the CAM-based seed area generated from the Transformer improves mIoU by 3.5\%, which reveals the Transformer is a more powerful alternative network for WSSS. Besides, our proposed two metrics verify that the Transformer can help the CAM-based seed area to alleviate under-activation by 11.0\% FNR, compared with the CNN. However, the CAM-based seed area generated from the Transformer also suffered from over-activation, which increases FPR by 5.8\%. Furthermore, we show some qualitative results of CAMs in Fig.~\ref{cnn_trans_features}. It reveals that CAMs generated from CNNs and Transformers are under- and over-activated, respectively.

\begin{figure*}[t]
	\centering
	%\fbox{\rule{0pt}{0.5in} \rule{0.9\linewidth}{0pt}}
	\footnotesize
	\begin{overpic}[width=0.98\linewidth]{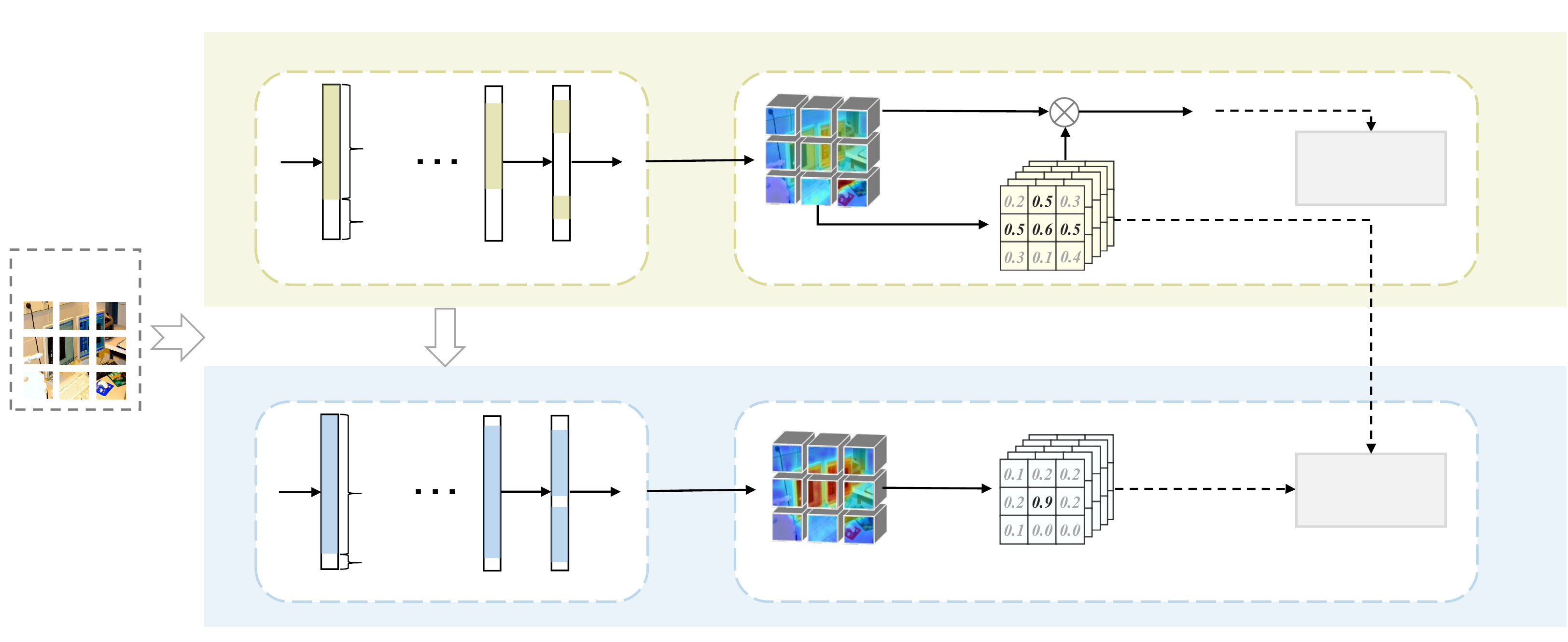}

        \put(83.3,10.2){\scriptsize Regularization}
        \put(85.5,8.5){\scriptsize$\mathcal{L}_{\text{REG}}$}
        \put(85.5,28.8){\scriptsize $\mathcal{L}_{\text{GEN}}$}
        \put(84.3,30.5){\scriptsize Generation }
        
        \put(93.6,22.3){ \textbf{Student}}
        \put(93.6,1.9){ \textbf{Teacher}}

        \put(69.2,34.2){\scriptsize{Normalize}}
        \put(66,31.5){{$\tau_{{1}}$}}

        \put(71.7,29.6){\scriptsize $\bm{\alpha}$}
        \put(70.6,5.9){\scriptsize $\bm{\tilde{\alpha}}$}

        \put(76.4,33.5){$\tilde{s}^c$}
        
        \put(19.3,24){ \scriptsize{layer$_{1}$}}
        \put(29.3,24){ \scriptsize{layer$_{{n\text{-}1}}$}}
        \put(34.3,24){ \scriptsize{layer$_{n}$}}
        
        \put(19.3,3.2){ \scriptsize{layer$_{1}$}}
        \put(29.3,3.2){ \scriptsize{layer$_{{n\text{-}1}}$}}
        \put(34.3,3.2){ \scriptsize{layer$_{n}$}}
        
        \put(62,4.5){ \scriptsize {Spatial Activation}}
        \put(63.5,3){ \scriptsize {Distribution}}

        \put(71.4,25){ \scriptsize {Spatial Activation}}
        \put(72.9,23.5){ \scriptsize {Distribution}}

        \put(57.2,10.2){\scriptsize {Softmax} }
        \put(57.2,27){\scriptsize {Softmax} }

        \put(49,25){ \scriptsize {Activation}}
        \put(50.3,23.5){ \scriptsize {{Value}}}
        \put(49,4.5){ \scriptsize {Activation}}
        \put(50.3,3){ \scriptsize {{Value}}}

        \put(30,19){ {\textbf{EMA}}}
        \put(57,37){ {\textbf{Activation Shape Controlling (Sec.~\ref{sec.4.1})}}}
        \put(56,16){ {\textbf{Activation Shape Regularzation~(Sec.~\ref{sec.4.2})}}}
        \put(21.5,37){ {\textbf{Strong  Adjustment}}}
        \put(21.5,16){ {\textbf{Weak  Adjustment}}}
        \put(2,23){ \textbf{Image}}
        \put(22.7,31){ \textcolor[RGB]{52,164,113}{\textit{pass}}}
        \put(22.7,26.4){ \textcolor[RGB]{236,95,116}{\textit{drop}}}
        \put(22.7,9){ \textcolor[RGB]{52,164,113}{\textit{pass}}}
        \put(22.7,4.7){ \textcolor[RGB]{236,95,116}{\textit{drop}}}

	\end{overpic}
	%\vspace{-1mm}
	\caption{The overall training process of our unified seed area generation paradigm (USAGE). The objective function of USAGE includes two terms: The first term is a generation loss which optimizes a mapping function between features and image-level labels controlled by the spatial activation distribution. The second term is a regularization loss to ensure the consistency between spatial activation distributions from two views. The two views are computed from a teacher network and a student network, respectively. The teacher and student networks are obtained by making different adjustments to the architecture of a classification network.}

	\label{fig:fig2}
\end{figure*}
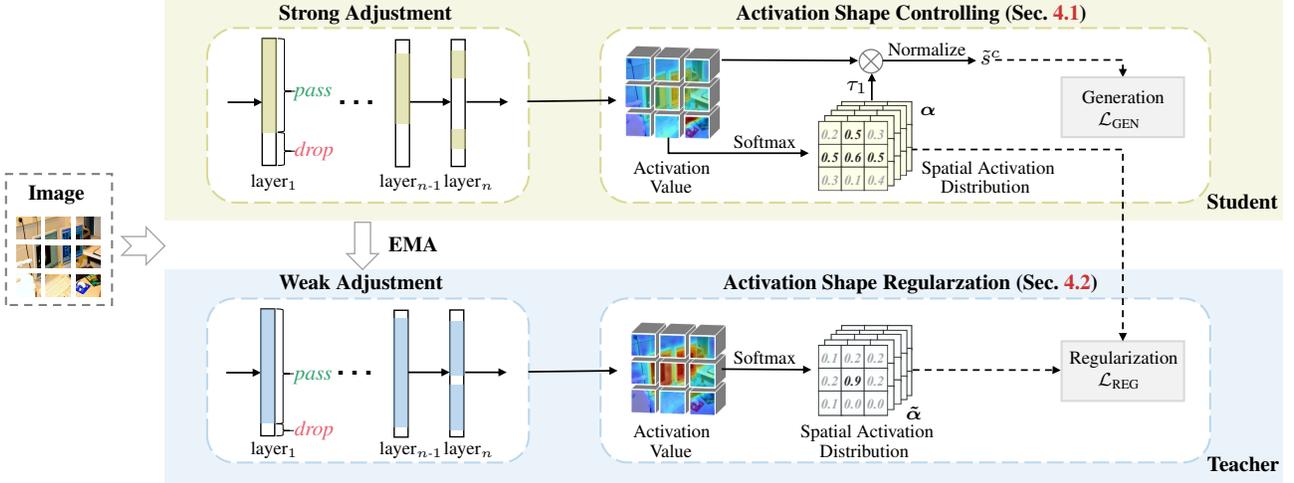

\subsection{Observation and Analysis}

We can observe that the feature characteristics of different networks would directly influence the shapes of CAM-based seed areas (over-activated or under-activated) by comparing the visualizations in both Fig.~\ref{fig:fig1} and Fig.~\ref{cnn_trans_features}. Specifically, the feature maps generated from the Transformer are more likely to involve more global context information. Meanwhile, the irrelevant information would not be suppressed from the following classifier, since the GAP follows the principle that each location on a feature map is capable of contributing to the classification results equally. Therefore, during inference, the seed area is unable to distinguish irrelevant context and the region of a target class, leading to the over-activation problem. In contrast, the CNN features capture local context, leading to under-activation. 

According to these experimental results, we observe that the core issue of the current seed area generation paradigm is that feature characteristics directly influence the shapes of seed areas. Thus, if we can design a mechanism to prevent the negative effects of the feature characteristics from both CNNs and Transformers, it would be capable of controlling the activation of seed areas in a unified paradigm.

\section{Unified Seed Area Generation}
In this section, we introduce the proposed \textbf{U}nified optimization paradigm for \textbf{S}eed \textbf{A}rea \textbf{GE}neration (USAGE). USAGE is motivated by previous analysis, in which the objective function consists of two terms: a generation loss $\mathcal{L}_{\text{GEN}} $ to make seed areas adapt to different types of networks and a regularization loss $\mathcal{L}_{\text{REG}}$ to overturn falsely activation seed areas, as shown in Fig.~\ref{fig:fig2}. The core of the generation loss is to introduce a \emph{spatial activation distribution} $\bm{\alpha}\in\mathbb{R}^{W\times H\times |\mathcal{C}|}$ to realize the shape of seed areas, which explicitly indicate the influence at each spatial location on the feature map contributed to the final classification result. By controlling the $\bm{\alpha}$ via smoothing or sharpening, the seed areas can be adaptive to the feature characteristics of different networks. Moreover, a regularization loss $\mathcal{L}_{\text{REG}} \left(\bm{\alpha}\right)$ is employed to regularize the seed areas.

The objective function of USAGE is formulated as:

\begin{equation}
\begin{aligned}
(\mathbf{\hat{A}}, \mathbf{\hat{w}}) &=\underset{\mathbf{A},\mathbf{w}}{\operatorname{argmin}}~\mathcal{L}_{\text{GEN}} \left(\mathbf{y}, \mathbf{\tilde{s}}\right) + \lambda \mathcal{L}_{\text{REG}} \left(\bm{\alpha}\right) \\
&=\underset{\mathbf{A},\mathbf{w}}{\operatorname{argmin}}~\frac{1}{|\mathcal{C}|} \sum_c\mathcal{L}_{\text{CE}}^c\left(y^c, \sigma\left(\tilde{s}^c\right)\right) + \lambda \mathcal{L}_{\text{REG}} \left(\bm{\alpha}\right),
\label{eq:USAGE_optimization}
\end{aligned}
\end{equation}
where $\lambda$ is a coefficient and 
\begin{equation}
\tilde{s}^c = \frac{\sum_{i=1}^W \sum_{j=1}^H \alpha_{i j}^c \sum_{d=1}^D w_d^c A_{i j}^d}{\sum_{i=1}^W \sum_{j=1}^H \alpha_{i j}^c}.
\label{temperature1}
\end{equation}
%where $\tau_1 \in \mathbb{R}_{+}$ denotes the temperature parameter~\cite{temp_cal_2017_pmlr}. 
Intuitively, the activation value $\sum_{d=1}^D w_d^c A_{i j}^d$ is an off-the-shelf prior for indicating the distribution of each location on the feature map during training. For simplicity, we employ it to compose the spatial activation distribution $\alpha_{i j}^c$. Moreover, we argue that $\alpha_{i j}^c$ should also be normalized along with the category channel since one location should only contribute to one class. In light of these clues, the spatial activation distribution $\bm{\alpha}$ is obtained by applying a softmax function to the activation value $\sum_{c=1}^D w_d^c A_{i j}^d$:
\begin{equation}
\alpha_{i j}^c = \frac{\exp \left(\sum_{d=1}^D w_d^c A_{i j}^d \right)}{\sum_{c=1}^{|\mathcal{C}|+1} \exp \left(\sum_{d=1}^D w_d^c A_{i j}^d \right)}. 
\end{equation}
We add an additional background channel with a constant value for rationality since not all locations are covered by foreground classes. 

\subsection{Activation Shape Controlling} 
\label{sec.4.1}
To cope with different types of networks, we integrate the temperature scaling~\cite{temp_cal_2017_pmlr} to control the spatial activation distribution. To this end, a simple modification to the mapping function (Eq.~\ref{temperature1}) is made as follows:
\begin{equation}
\tilde{s}^c = \frac{\sum_{i=1}^W \sum_{j=1}^H (\alpha_{i j}^c)^{1/ \tau_{1}} \sum_{d=1}^D w_d^c A_{i j}^d}{\sum_{i=1}^W \sum_{j=1}^H (\alpha_{i j}^c)^{1/ \tau_{1}}},
\label{temperature1.1}
\end{equation}
where $\tau_1 \in \mathbb{R}_{+}$ denotes a temperature parameter, it can control spatial activation distribution to be sharpened or smoothed, \emph{e.g.,} a smaller value of $\tau_1$ encourages the network to focus on a portion of locations with higher contribution and alleviate over-activation for Transformers. In contrast, when $\tau_1$ is nearly infinite, it will force the network to learn from features at each location, which is beneficial for CNNs.

\begin{figure}[t]
	\centering
	%\fbox{\rule{0pt}{0.5in} \rule{0.9\linewidth}{0pt}}
	\footnotesize
	\begin{overpic}[width=0.98\linewidth]{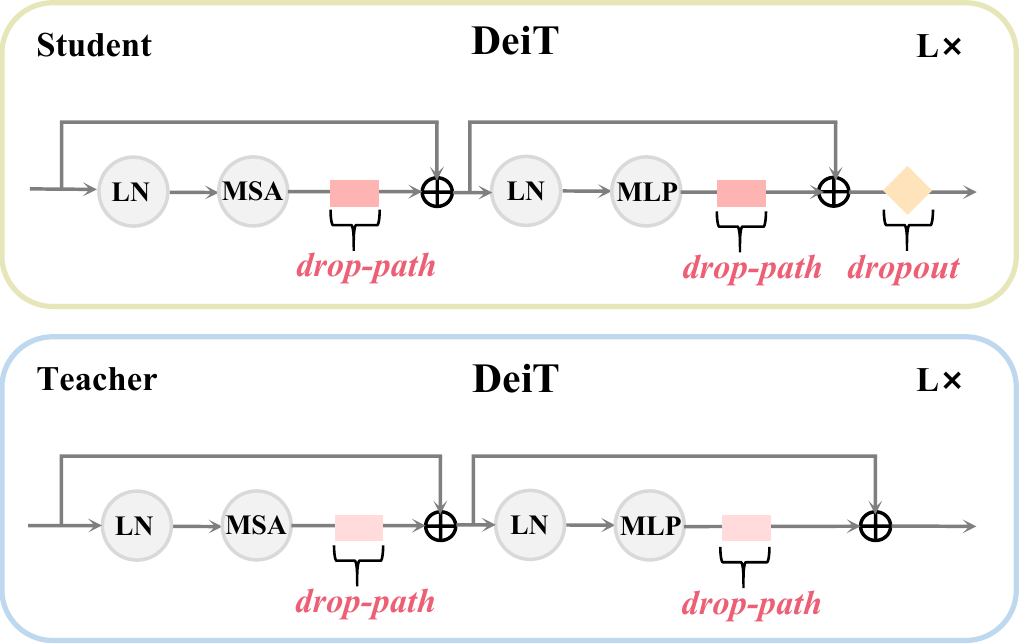}
	\end{overpic}
	%\vspace{-1mm}
  \vskip 0.05in
	\caption{The diagram of how to make adjustment to the architecture of a classification network (DeiT~\cite{Trans_tde_2021_ICML}) to form the student network and the teacher network. Two adjustment operation are used, \emph{i.e.,} dropout~\cite{Drop_drop_out_2014_JMLR} and drop-path~\cite{Drop_drop_path_2016_arxiv}. } 
	%\vspace{-4mm}
 
	\label{fig:fig3}
\end{figure}

\subsection{Activation Shape Regularization} 
\label{sec.4.2}
Since the spatial activation distribution $\bm{\alpha}$ is highly correlated with learned features, it lacks the ability to rectify mistakes caused by feature representations \emph{i.e.,} overturning incorrectly activated seed areas. To solve this problem, we apply a regularization loss to regularize the spatial activation distribution to encourage semantic consistency between two views.

\noindent{\textbf{Self-adaptive Network Adjustment.}} The different views here are obtained by making different adjustments to the architecture of the classification network, where the adjustment is achieved by applying dropout~\cite{Drop_drop_out_2014_JMLR} and drop-path~\cite{Drop_drop_path_2016_arxiv} to the architecture (the detailed design is shown in Fig.~\ref{fig:fig3} as an example). Specifically, we adopt a strong adjustment and a weak adjustment to the architecture, resulting in a student network and a teacher network, respectively. The weights of the teacher are updated from the student using Exponential Moving Average (EMA)~\cite{EMA_EMA_1977_CUP}.  However, maintaining a fixed adjustment gap between the teacher and student models may overlook the varying learning difficulties across different stages of training. To address this, we propose a self-adaptive network adjustment scheme that can dynamically increase the adjustment gap based on the current learning status of the models. We estimate the current learning status as the EMA of the classification score at each training time step, as inspired by recent work~\cite{freematch_2022_arxiv}. We use the cross entropy loss function to measure the consistency between the spatial activation distribution $\bm{\alpha}_{i j}$ from the student network and the spatial activation distribution $\tilde{\bm{\alpha}_{i j}}$ from the teacher network. The regularization term is specified as:

\begin{equation}
	\mathcal{L}_{\text{REG}} \left(\bm{\alpha}\right) =-\tau^2_{2} \frac{1}{N} \sum_{i=1}^W \sum_{j=1}^H (\tilde{\bm{\alpha}_{i j}})^{1 / \tau_{2}} \log \left(\bm{\alpha}_{i j}\right)^{1 / \tau_{2}},
\label{USAGE_REG}
\end{equation}
where $\tau_{2} \in \mathbb{R}_{+}$ is a temperature parameter.%~\cite{temp_cal_2017_pmlr}. 

\subsection{Relationship to Other Variants} 
\label{sec.4.3}
In this subsection, we analyze the relationship between USAGE and other seed area generation approaches, including cross-view regularization by geometric transform, multiple instance learning and class-to-patch attention mapping. We show that all of them are special cases of USAGE and conduct throughout comparison with them in Sec.~\ref{sec:abla}.

\noindent\textbf{Cross-view Regularization by geometric Transform.} 
Most previous WSSS methods~\cite{WSSS_L2G_2022_CVPR,WSSS_SEAM_2020_CVPR} cross-view regularization by geometric transform, \emph{i.e.}, applying different manually pre-defined geometric transforms to input images to construct multiple views. Their objective function for seed area generation has the same form as ours, \emph{i.e.}, Eq.~\ref{eq:USAGE_optimization}, but they constructed the mapping of the classifier $\mathcal{G}(\cdot,\cdot)$ by Eq.~\ref{eq:1} rather than Eq.~\ref{temperature1.1}. Without the activation shape controlling ability provided by Eq.~\ref{temperature1.1}, their methods are easily influenced by the feature characteristics of different networks, and thus suffer from problematic activations. Besides, we also show that our instantiation for regularization, \emph{i.e.}, cross-view regularization by architecture adjustment is more effective for seed area generation from Transformers (Sec.~\ref{sec:abla}).

\noindent\textbf{Multiple Instance Learning.}
In the early stage, \emph{i.e.,} prior to the popularization of the CAM, pioneer WSSS methods used multiple instance learning (MIL)~\cite{MIL_framework_1997_NIPS} to mine seed areas. For example, Pinheiro \emph{et al.}~\cite{WSSS_FIP_2015_CVPR} instantiated the mapping of the classifier $\mathcal{G}(\cdot,\cdot)$ as a MIL problem by leveraging Eq.~\ref{temperature1}, and computed the objective function following Eq.~\ref{baseline_objective}. However, since Eq.~\ref{temperature1} lacks activation shape controlling as well as Eq.~\ref{baseline_objective} lacks the regularization term, MIL-based methods are easily affected by different feature characteristics, which leads to problematic activations.

\noindent\textbf{Class-to-patch Attention Mapping.}
As mentioned in Sec.~\ref{sec:related}, Xu \emph{et al.}~\cite{WSSS_MCT_2022_CVPR} proposed to generate seed areas from Transformers based on the class-to-patch attention map rather than the CAM~\cite{CAM_cam_2018_CVPR}. They used Eq.~\ref{baseline_objective} as the objective function for seed area generation and instantiated the mapping of the classifier $\mathcal{G}(\cdot,\cdot)$ by:
\begin{equation}
\label{eq:6}
\hat{s}^c = \frac{\sum_{i=1}^W \sum_{j=1}^H \beta_{i,j}^c \sum_{d=1}^D  \frac{1}{D} \mathcal{P}(A_{i j}^d)}{\sum_{i=1}^W \sum_{j=1}^H \beta_{i,j}^c},
\end{equation}
where
\begin{equation}
\beta_{i,j}^c = w_d^c A_{i, j}^d,
\end{equation}
and $\mathcal{P(\cdot)}$ is a multilayer perceptron (MLP) layer. Eq.~\ref{eq:6} shows that \cite{WSSS_MCT_2022_CVPR} directly used the activation value to indicate the influence at each spatial location rather than a normalized distribution. 
Since they neither explicitly controlled the shape of activation (Eq.~\ref{eq:6}) nor involved the regularization term (Eq.~\ref{baseline_objective}), the seed areas generated from Transformers are inevitably over-activated (see the second column of Fig.~\ref{fig:fig6}).

%% file: tab/comp_cnn_trans.tex
\setlength{\columnsep}{8pt}%
\begin{table}[!tp]
	\newcommand{\CC}[1]{\cellcolor{gray!#1}}

	\centering
	\small
\resizebox{0.98\linewidth}{!}{
	\begin{tabular*}{8.7cm}{@{\extracolsep{\fill}}c|ccc}
		\toprule
		Methods & mIoU~(\%)$\uparrow$ & FPR~(\%)$\downarrow$ & FNR~(\%)$\downarrow$    \\ \midrule
	    CNN.~CAM      &   47.8    &    22.6   &    30.1       \\
        CNN.~USAGE w/o REG	& 49.6      &  22.9   &    27.2   \\
		CNN.~USAGE      & 57.7    &    20.4    &    22.6    \\ \midrule
  	Trans.~CAM      &  52.3     &   28.4     &  19.1      \\
        Trans.~USAGE w/o REG	    & 64.0   &   20.3    &  15.5                 \\
		Trans.~USAGE	    &\textbf{67.7}   &\textbf{17.8}   &\textbf{14.7}           \\
        %\cmark           &  \cmark   &    \cmark         &  \CC{15}\textbf{66.6}         \\
        %\cmark           &  \cmark     &    \cmark         & \textbf{66.0}          \\
 \bottomrule
	\end{tabular*}}
	%\vspace{-4mm}
 \vskip 0.1in
   	\caption{Analysis of different seed areas computed from the CNN and the Transformer on the PASCAL VOC 2012 \textit{train} set~\cite{dataset_pascal_2010_IJCV}.} 
%Baseline: the transformer-based MIL framework we proposed while w/o self-knowledge distillation. Student branch: The strong regularization is applied. Teacher branch: Using extra branch to provide additional supervision.}
	\vspace{-3mm}
	\label{tab_cnn_trans}
\end{table}

%% file: sec/4_results.tex
\section{Experiments}
In this section, we first describe the experimental setup and implementation details (Sec.~\ref{sec:ex_se}). Then, we compare our method with state-of-the-art weakly supervised semantic segmentation methods (Sec.~\ref{sec:sota}). Finally, we conduct an ablation study to show the contribution of each component in our method (Sec.~\ref{sec:abla}).
\subsection{Experimental Settings}
\label{sec:ex_se}
\par\noindent\textbf{Datasets.} We evaluate the proposed approach on two datasets, \ie, PASCAL VOC 2012~\cite{dataset_pascal_2010_IJCV} and MS COCO 2014~\cite{dataset_coco_2014_ECCV}. \textbf{PASCAL VOC} has three subsets, \ie, training (train),  validation (val) and test sets, each of which contains 1,464, 1,449, and 1,456 images, respectively. It has 20 object classes and one background class for the semantic segmentation task. Following~\cite{WSSS_ficklenet_2019_CVPR,WSSS_CP_2021_ICCV,WSSS_L2G_2022_CVPR,WSSS_causal_2020_NIPS}, an augmented set of 10,582 images, with additional data from~\cite{dataset_SBD_2011_ICCV}, is used for training. \textbf{MS COCO} contains 80 object classes and one background class for semantic segmentation. Its training and validation sets contain 80K and 40K images, respectively.

\input{tab/abla_pgt_voc}

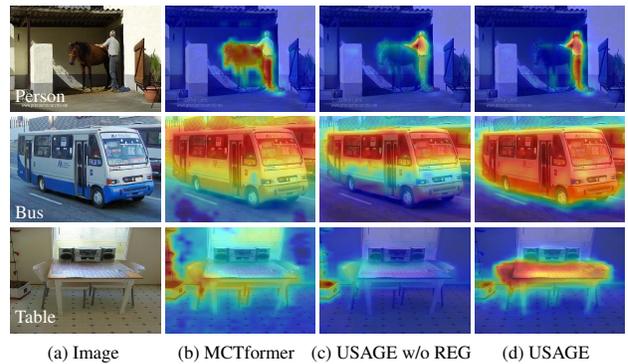
\begin{figure}[t]
	\centering
        \scriptsize
	\begin{overpic}[width=0.98\linewidth]{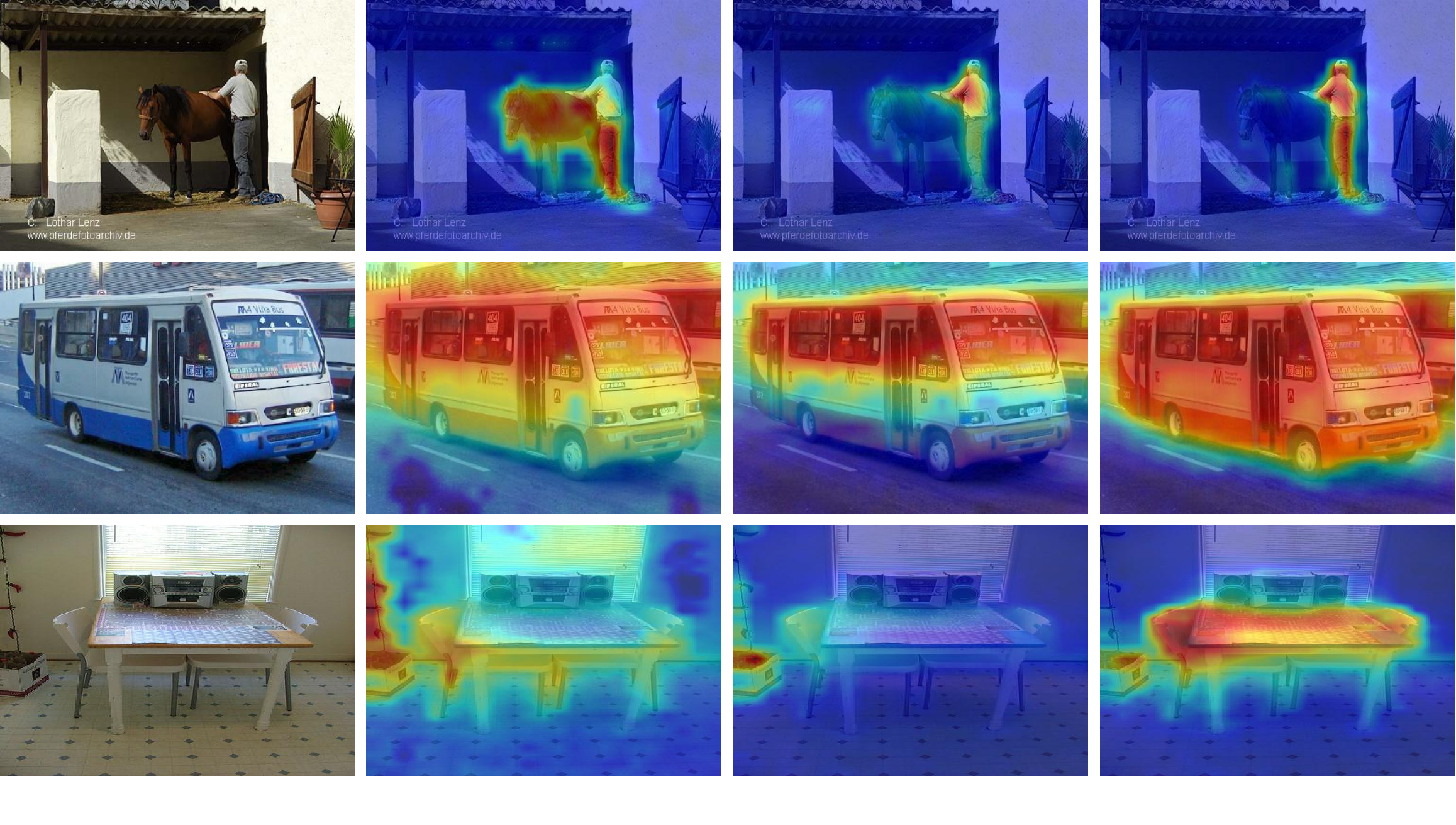}
            \put(0.7,5){\textcolor{white}{Table}}
            \put(0.7,22){\textcolor{white}{Bus}}
            \put(0.7,41){\textcolor{white}{Person}}
		\put(6,-0.8){(a) Image}
		\put(27.2,-0.8){(b) MCTformer}
		\put(48.9,-0.8){(c) USAGE~w/o REG}
		\put(80,-0.8){(d) USAGE}
	\end{overpic}
 \vskip 0.05in
	%\vspace{-1mm}
	\caption{Seed area visualization. (a) Input image. (b) Seed area generated in MCTformer~\cite{WSSS_MCT_2022_CVPR}. (c) USAGE (Ours) without using activation shape regularization. (d) USAGE (Ours).} %``Trans. MIL'' denotes transformer-based MIL.}
 \label{fig:fig6}
 
\end{figure}

\par\noindent\textbf{Evaluation metrics.}
Following prior works~\cite{WSSS_MCT_2022_CVPR}, we use the mIoU to evaluate 
the semantic segmentation performance on the \textit{val} set, of the two benchmarks. We obtained the semantic segmentation results on the PASCAL VOC \textit{test} set from the official evaluation server.
\par\noindent\textbf{Implementation details.}
We perform our USAGE on two types of networks, \emph{i.e.,} Transformer and CNN. For Transformer, following~\cite{WSSS_MCT_2022_CVPR}, we use the DeiT-S~\cite{Trans_tde_2021_ICML} pre-trained on ImageNet~\cite{dataset_imagenet_2009_cvpr} as our network. We followed the data augmentation and default training parameters provided in~\cite{Trans_tde_2021_ICML}. Besides, we also employ the patch-level pairwise affinity proposed by~\cite{WSSS_MCT_2022_CVPR} to refine the seed area without additional computations. We set $\tau_1$ to 1 and $\tau_2$ to 0.1. We set the drop path rate $\gamma_t$ in the teacher network to 0.05. The drop path rate $\gamma_s$ and the drop rate $\delta_s$ in the student network are raised from 0.05 to 0.15 and from 0 to 0.01, respectively. The $\lambda$ is 0.25 and the update rate for EMA is set to 0.99. For CNN, we followed the procedure of~\cite{WSSS_psa_2018_CVPR}, including the use of ResNet38~\cite{SS_ResNet38_2019_PR}. We set $\tau_1$ to 50 as a way of smoothing, and $\tau_2$ also to 0.1. To report final semantic segmentation results, we follow prior works~\cite{WSSS_psa_2018_CVPR,WSSS_CP_2021_ICCV,WSSS_MCT_2022_CVPR} to use ResNet38~\cite{SS_ResNet38_2019_PR} and ResNet101~\cite{network_resnet50_2016_CVPR} as the backbones for segmentation. Notably, since the seed area computed from the Transformer achieves better performance, we adopt DeiT-S~\cite{Trans_tde_2021_ICML} as our default network for seed area generation.

\input{tab/sota_voc}

\input{tab/sota_coco}

\subsection{Comparisons with the State-of-the-Arts}
\label{sec:sota}
\begin{figure}[t]
	\centering
	%\fbox{\rule{0pt}{0.5in} \rule{0.9\linewidth}{0pt}}
	\footnotesize
	\begin{overpic}[width=0.98\linewidth]{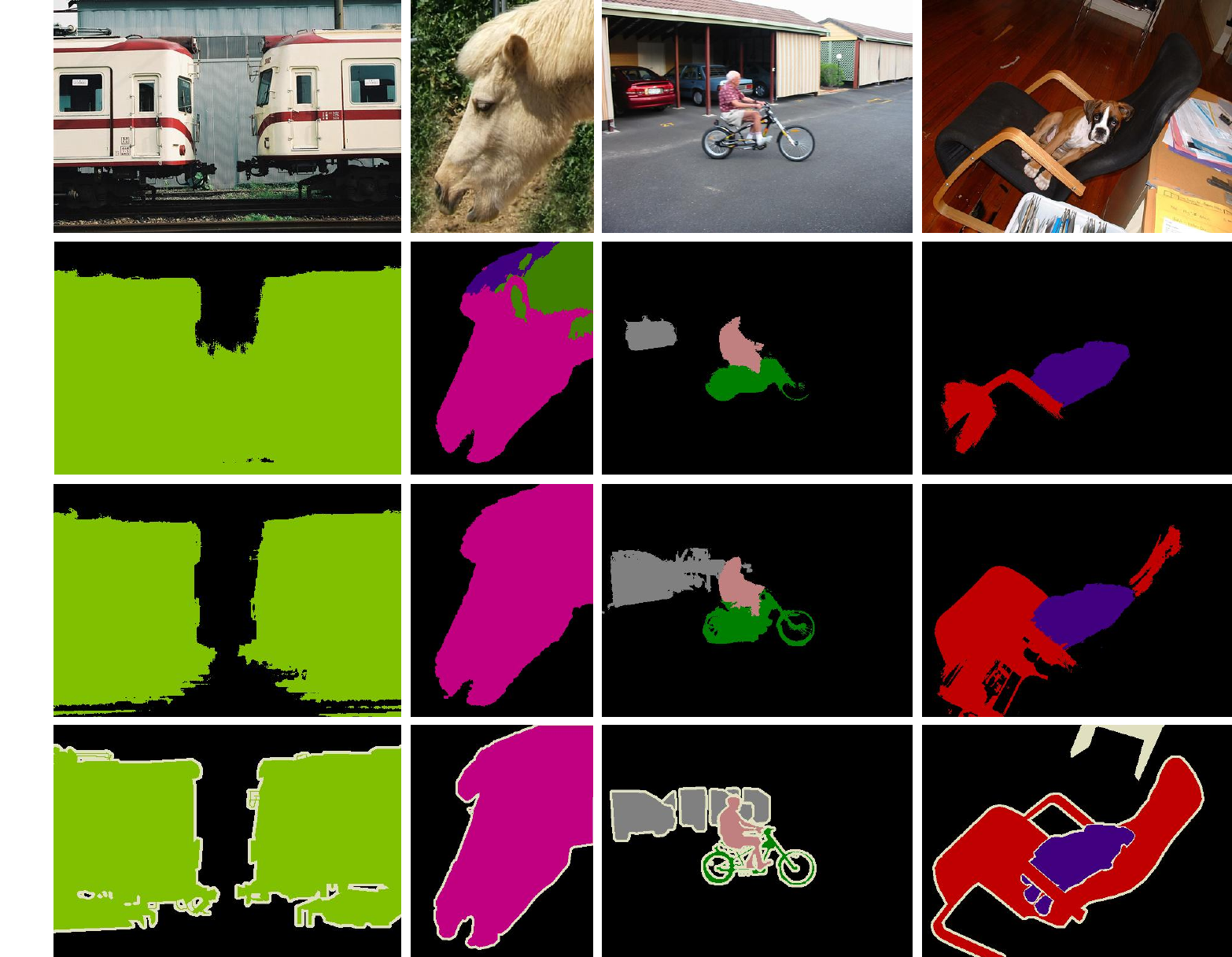}
        \put(-0.5,67){(a)}
        \put(-0.5,47.5){(b)}
        \put(-0.5,29){(c)}
        \put(-0.5,9){(d)} 
	\end{overpic}
	%\vspace{-1mm}
 \vskip 0.05in
	\caption{ Qualitative segmentation results on the PASCAL VOC 2012~\cite{dataset_pascal_2010_IJCV} \emph{val} set. (a) Input image. (b) MCTformer~\cite{WSSS_MCT_2022_CVPR}. (c) USAGE (Ours). (d) Ground-truth.}
 %\vspace{-1pt}
 \label{fig:sota_res38}
\end{figure}
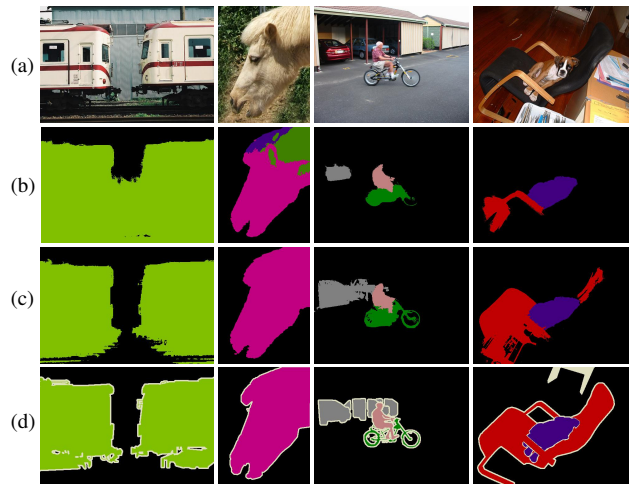

\noindent\textbf{PASCAL VOC.} We follow~\cite{WSSS_psa_2018_CVPR,WSSS_SEAM_2020_CVPR,WSSS_MCT_2022_CVPR} to apply PSA~\cite{WSSS_psa_2018_CVPR} on the proposed seed areas (seed) to generate pseudo masks (mask) on the train set. As shown in Table~\ref{tab:seed}, the proposed USAGE performs better than existing works by large margins in terms of seed area generation and pseudo mask generation, \emph{e.g.,} improving previous SOTA~\cite{WSSS_max_2022_ECCV} by 4.1\%.  Table~\ref{tab:sota_res38} shows that the proposed \nn~achieves segmentation results (mIoUs) of 71.9\% and 72.8\% on the val and test sets, respectively. The proposed \nn~performs significantly better than all the existing methods using only image-level labels. We also note that, our results achieve similar performance with most WSSS methods which leverage saliency maps. We also show some qualitative results of the seed areas in Fig.~\ref{fig:fig6} and segmentation results in Fig.~\ref{fig:sota_res38}. This indicates our USAGE is effective to generate high-quality results. Even if the context is ambiguous,  our method still performs well.

%we present some examples of the final semantic segmentation results, where our results are close to the ground truthed segmentation.

\noindent\textbf{MS COCO.} Table~\ref{tab:segsota_coco} shows the comparison results of our method against several existing methods on the MS COCO 2014 dataset~\cite{dataset_coco_2014_ECCV}. For example, by adopting the ResNet101 as the segmentation network backbone, our approach surpasses previous methods using only image-level labels by 2.3\%. In particular, our method can even achieve comparable results compared to the methods using additional saliency maps. We also offer a superior performance using the ResNet38 as the segmentation network backbone, achieving 42.7\% mIoU.  More qualitative results are shown in the supplementary material.

\subsection{Ablation Study}
\label{sec:abla}

\noindent\textbf{Ablation of main components.}  
Here, we do an ablation study to show the benefit brought by each component of our proposed~\nn, \emph{i.e.,} activation shape controlling and activation shape regularization. As shown in Table~\ref{tab_cnn_trans}, by introducing the activation shape controlling on the spatial activation distribution $\bm{\alpha}$, our method can obtain much better mIoUs for the seed area computed from the CNN and the Transformer (49.6\% and 64.0\%). Meanwhile, the activation shape controlling yields great performance improvements on the other two metrics, FPR and FNR, by 3.9\% and 3.3\% on average. This reveals that the activation shape controlling alleviates problematic activations caused by feature characteristics. Furthermore, the seed area results are further boosted to 57.7\% and 67.7\%, respectively, after integrating the activation shape regularization. As shown in the Fig.~\ref{fig:fig6}, we observe that the seed areas appear more complete (the first two rows) and accurate (the last row), which verifies the ability of the activation shape regularization in overturning misled seed areas.

\input{tab/smooth_sharpen}

\noindent\textbf{Discussion of temperature parameter.} We conduct experiments of the hyper-parameter temperature $\tau_{1}$, as shown
in Table~\ref{tab_smooth_sharpen}. The results show that controlling temperature $\tau_{1}$ to adapt to different types of networks is easy, as it follows a deterministic principle, \emph{i.e.,} smoothing for CNNs or sharpening for Transformers, and the performance is robust across different temperature values.

\noindent\textbf{Rates of drop path and dropout for our network adjustment.} 
We use drop path and dropout to realize network adjustment, where whether an adjustment is strong or weak is determined by the rates of both drop path and dropout. High and low rates lead to strong and weak adjustments, respectively.  If the difference between the rates for the teacher and that for the student increases, then their adjustment gap becomes larger. The ``weak/strong adjustment'' encourages a model to produce consistent predictions for the same pixel under different views, leading to robust localization. As shown in Table.~\ref{abla_rate4}, our USAGE demonstrates robustness \emph{w.r.t.} these hyper-parameters. $\gamma_t$: drop path rate in the teacher model (default setting). $\gamma_s$: drop path rate in the student model. $\delta_s$: drop rate in the student model.

\noindent\textbf{Discussion of Self-adaptive strategy for our network adjustment.}
In order to control the growing gap between the student and teacher's network adjustment, we introduce a self-adaptive strategy. By adaptively increasing the adjustment gap, the student will efficiently learn to distinguish and preserve all features that are pertinent to the foreground objects. We demonstrate the effectiveness of the self-adaptive strategy by comparing it with two commonly used functions, namely ``Fixed'', maintaining the adjustment gap at a fixed value, and ``Linear'', an increase in the adjustment gap at a fixed value in a linear fashion with training progresses. As shown in Table~\ref{ab_sa}, our proposed strategy achieves better performance with ``Fixed'' and ``Linear'' strategies, which indicates that the self-adaptive strategy is essential for the proposed network adjustment framework.

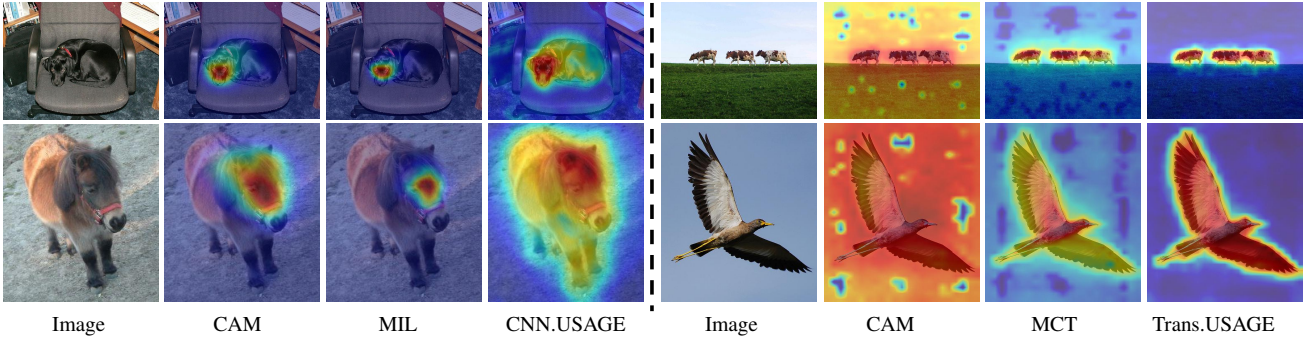
\begin{figure*}[t]
	\centering
	%\fbox{\rule{0pt}{0.5in} \rule{0.9\linewidth}{0pt}}
	\footnotesize
	\begin{overpic}[width=0.99\linewidth]{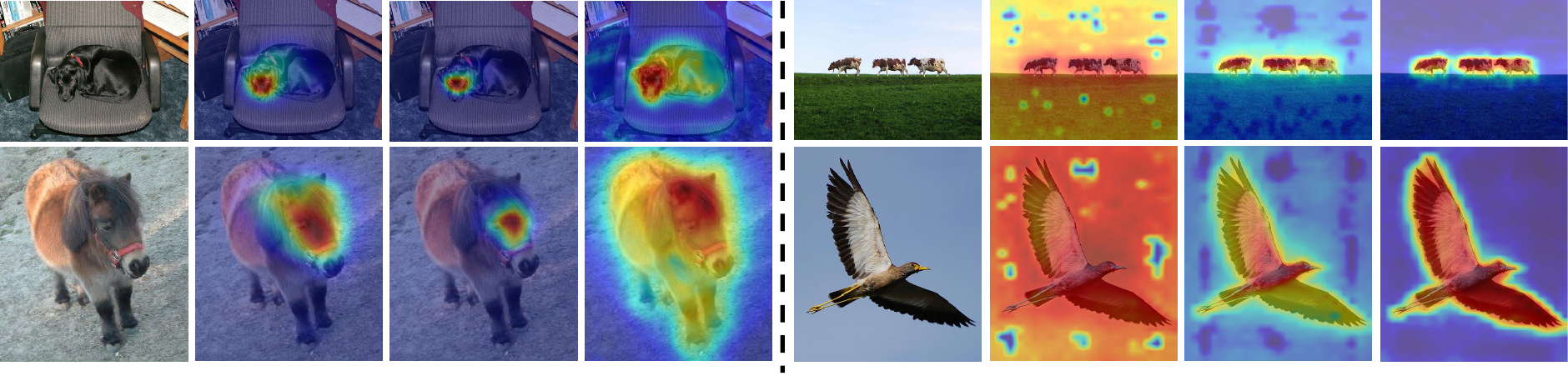}
		\put(3.8,-1){Image}
		\put(16.2,-1){CAM}
  	\put(28.9,-1){MIL}
		\put(38.7,-1){CNN.USAGE}
        \put(54,-1){Image}
  	\put(66.4,-1){CAM}
		\put(79.1,-1){MCT}
		\put(88.4,-1){Trans.USAGE}
		%\put(80,3){(d) USAGE}
	\end{overpic}
	\vspace{10pt}
	\caption{Seed area visualizations on the PASCAL VOC~\cite{dataset_pascal_2010_IJCV} \emph{train} set. Left: Seed area generation from a CNN (ResNet38~\cite{SS_ResNet38_2019_PR}). Right: Seed area generation from a Transformer (DeiT-S~\cite{Trans_tde_2021_ICML}). MCT: Class-to-patch attention map~\cite{WSSS_MCT_2022_CVPR}. MIL: a MIL-based method~\cite{WSSS_FIP_2015_CVPR}.}
%\vspace{10pt}
 \label{fig:seed2}
\end{figure*}

\noindent\textbf{Variants.} The objective function of our USAGE (Eq.~\ref{eq:USAGE_optimization}) includes a classification term and a regularization term. The first term aims to optimize a mapping function, which is realized by the activation shape controlling (CM) in our USAGE according to Eq.~\ref{temperature1.1}. As discussed in Sec.~\ref{sec.4.3}, the mapping function can also be realized by three variants, including CAM~\cite{CAM_cam_2018_CVPR}, class-to-patch attention mapping~\cite{WSSS_MCT_2022_CVPR} and MIL-based method~\cite{WSSS_FIP_2015_CVPR}. 
The second term is achieved by network architecture adjustment (AA), which also can be achieved by two variants, \emph{i.e.,} geometric transform~\cite{WSSS_SEAM_2020_CVPR} and online attention accumulation~\cite{WSSS_OAA_2019_ICCV}. As shown in Table~\ref{tab:abla_variants}, we conduct experiments to evaluate these variants, which reveals our terms are able to achieve superior results in terms of seed area generation from both CNNs and Transformers, and performs favourably against all variants. Based on the experiment results, we have the following observations: 1) Compared to CNN, we argue that Transformer is a more effective network for seed area generation (52.3\% vs. 48.7\%), especially when using MIL as a type of mapping (54.3\% vs. 29.3\%). 2) We replace our network adjustment with geometric transform~\cite{WSSS_SEAM_2020_CVPR} for regularization and its performance dramatically drops by 5.3\%. We analyze that the reason might be the regularization by geometric transform barely overturn misled seed areas, since erroneous seed areas are also capable of keeping consistent between different transformations. Online attention accumulation~\cite{WSSS_OAA_2019_ICCV} also shows inferior performance, with performance drops by 3.6\% on average. 3) Even without using our mapping (\emph{i.e.,} CM), our AA can also bring performance gains for seed area generation (56.5\% vs. 52.3\%), which reveals its ability to overturn misled seed areas. We show a few qualitative results comparing seed areas among different variants in Fig.~\ref{fig:seed2}. More comparisons are provided in the supplementary material.

\input{tab/abla_real4}
\input{tab/abla_sa}

\input{tab/abla_variants}

%% file: tab/abla_pgt_voc.tex
% \begin{table}
% \centering
% \resizebox{\linewidth}{!}{ %< auto-adjusts font size to fill line
% \begin{tabular}{@{}lccc@{}}
% \toprule
% Method & Frobnability & Frobnability & Frobnability \\
% \midrule
% Ours v0 & Frumpy \\
% Ours v1 & Frobbly \\
% \textbf{Ours} & Makes one's heart Frob\\
% \bottomrule
% \end{tabular}
% } %< \resizebox
% \caption{
% % 
% \textbf{Ablations} -- our decisions are well justified.
% % 
% } % \caption
% \label{tab:ablations}
% \end{table}

\begin{table}[t]
\renewcommand{\arraystretch}{1.2}
\small
\centering
\resizebox{1.0\linewidth}{!}{
\begin{tabular*}{9cm}{@{\extracolsep{\fill}}lcc}
\toprule[1pt] 
Method            & Seed & Mask                   \\ \midrule
%PSA (CVPR18) \cite{kolesnikov2016seed} & 48.0 & 61.0 \\
% Mixup-CAM (BMVC20) \cite{} &50.1&61.9 \\
Chang~\emph{et al.} (CVPR20) \cite{WSSS_SCE_2020_CVPR} & 50.9&63.4 \\
SEAM (CVPR20)~\cite{WSSS_SEAM_2020_CVPR} &55.4& 63.6\\
AdvCAM (CVPR21)~\cite{WSSS_AdvCAM_2021_CVPR} &55.6& 68.0 \\
CDA (ICCV21)~\cite{WSSS_CDA_2021_ICCV} &58.4&66.4 \\
Zhang~\emph{et al.}(ICCV21)~\cite{WSSS_CP_2021_ICCV} &57.4& 67.8 \\
RIB (NeurIPS21)~\cite{WSSS_rib_2021_NIPS} &56.5& 68.6 \\
SIPE (CVPR22)~\cite{WSSS_SIPE_2022_CVPR} &58.6&  - \\
ReCAM (CVPR22)~\cite{WSSS_Recam_2022_CVPR}  &56.6& - \\
CLIMS (CVPR22)~\cite{WSSS_CLIMS_2022_CVPR} &56.6& 70.5 \\
MCTformer (CVPR22)~\cite{WSSS_MCT_2022_CVPR}  &61.7& 69.1 \\
ViT-PCM (ECCV22)~\cite{WSSS_max_2022_ECCV} &63.6 & 67.1 \\

% DSRG (CVPR18) \cite{huang2018weakly} & VGG16 & I+S&26.0 \\
% Wang \etal (IJCV20) \cite{wang2020weakly}&VGG16&I&27.7\\
%     Luo \etal (AAAI20)  \cite{luo2020learning}  &     VGG16              &  I&    29.9 \\
% SEAM (CVPR20) \cite{wang2020self} & ResNet38 & I&31.9 \\
% CONTA (NeurIPS20) \cite{zhang2020causal} & ResNet38 & I&32.8 \\
% EPS (CVPR21) \cite{lee2021railroad} & ResNet101 & I+S & 35.7 \\
% AuxSegNet (ICCV21) \cite{xu2021leveraging} & ResNet38 & I+S & 33.9 \\
% Kweon \etal (ICCV21) \cite{kweon2021unlocking} & ResNet38 & I & 36.4 \\
% CDA (ICCV21) \cite{su2021context} & ResNet38 & I &33.2 \\
 \midrule
\textbf{\nn} (Ours) & \textbf{67.7} & \textbf{72.8} \\
  \bottomrule[1pt]
\end{tabular*}
}
\vskip 0.1in
\caption{Evaluation of the seed area (Seed) and the corresponding pseudo mask (Mask) refined by PSA \cite{WSSS_psa_2018_CVPR} in terms of mIoU (\%) on the PASCAL VOC 2012~\cite{dataset_pascal_2010_IJCV} \textit{train} set. }
\label{tab:seed}
%\vspace{-4mm}
% \end{center}
%\vspace{-18pt}
\end{table}

%% file: tab/sota_voc.tex
\begin{table}[t]
\centering
% \begin{center}
\small
% \footnotesize
\resizebox{1.0\linewidth}{!}{
\renewcommand{\arraystretch}{1.2}
\begin{tabular*}{8.9cm}{@{\extracolsep{\fill}}lcccc}

% \begin{tabularx}{0.48\textwidth}{p{0.46\linewidth}p{0.13\linewidth}c{0.01\linewidth}p{0.01\linewidth}p{0.01\linewidth}}
% \begin{tabular}{lccc}
\toprule[1pt]
 Method            & Seg.Back. & Sup.          & Val            & Test           \\ \midrule
SeeNet (NeurIPS18) \cite{WSSS_SE_2018_NIPS} & ResNet38 & I+S &63.1 & 62.8 \\
Sun \etal (ECCV20)~\cite{WSSS_MCIS_2020_ECCV}   &          ResNet101        &I+S&        66.2          &      66.9        \\
 EPS (CVPR21)~\cite{WSSS_railroad_2021_CVPR} & ResNet101 & I+S & 70.9&70.8 \\
 AuxSegNet (ICCV21)~\cite{WSSS_LAT_2021_ICCV} & ResNet38 & I+S &69.0&68.6 \\ 
 %DRS (AAAI21) \cite{} & ResNet101 & I+S &70.4 & 70.7 \\
 ESOL (NeurIPS22)~\cite{WSSS_expand_sh_2022_NIPS} & ResNet101 &I+S &71.1 & 70.4 \\
 ReCAM (CVPR22)~\cite{WSSS_AdvCAM_2021_CVPR} & ResNet101 & I+S &71.8&72.2 \\
 L2G (CVPR22)~\cite{WSSS_L2G_2022_CVPR} & ResNet101 & \textbf{I+S} & \textbf{72.1}& \textbf{71.7} \\ \hline
 %RCA (CVPR22)~\cite{WSSS_PPC_2022_CVPR} & ResNet101 & I+S & \textbf{72.2}& \textbf{72.8} \\ \hline
Araslanov \etal (CVPR20) \cite{WSSS_SSSS_2020_CVPR}    &          ResNet38    &I     &      62.7        &   64.3       \\ 
SEAM (CVPR20) \cite{WSSS_SEAM_2020_CVPR} & ResNet38 &I& 64.5 & 65.7 \\
 BES (ECCV20) \cite{WSSS_be_2020_ECCV} & ResNet101 & I &65.7&66.6 \\
 CONTA (NeurIPS20) \cite{WSSS_causal_2020_NIPS}   &          ResNet38         &  I&    66.1       &    66.7            \\  
AdvCAM (CVPR21) \cite{WSSS_AdvCAM_2021_CVPR} & ResNet101 & I &68.1 &68.0 \\
 ECS-Net (ICCV21) \cite{WSSS_ECS_2021_ICCV} & ResNet38 & I &66.6 & 67.6 \\
 CDA (ICCV21) \cite{WSSS_CDA_2021_ICCV} & ResNet38 & I &66.1&66.8 \\
 Zhang \etal (ICCV21) \cite{WSSS_CP_2021_ICCV} & ResNet38 & I & 67.8&68.5 \\
AdvCAM (CVPR21) \cite{WSSS_AdvCAM_2021_CVPR} & ResNet101 & I &68.1 &68.0 \\
RIB (NeurIPS21) \cite{WSSS_rib_2021_NIPS} & ResNet38 & I &68.3 & 68.6 \\
 SIPE (CVPR22) \cite{WSSS_SIPE_2022_CVPR} & ResNet38 & I& 68.8& 69.7 \\ 
 CLIMS (CVPR22) \cite{WSSS_CLIMS_2022_CVPR} & ResNet101 & I & 70.4 & 70.0 \\
 MCTformer (CVPR22) \cite{WSSS_MCT_2022_CVPR}& ResNet38 & I & ~71.1$^*$ & 71.6 \\
 \midrule
%   \textbf{\nn}$^{*}$ (Ours) & ResNet38 & I&   66.1 &  66.1 \\
  \textbf{\nn} (Ours) & ResNet38 &\textbf{I}&    \textbf{71.9} &  \textbf{72.8}\\
  \bottomrule[1pt]
% \end{tabularx}
\end{tabular*}
}
% \end{center}
%\vspace{-17pt}
\vskip 0.1in
\caption{Performance comparison of WSSS methods in terms of mIoU (\%) on the PASCAL VOC 2012~\cite{dataset_pascal_2010_IJCV} \textit{val} and \textit{test} sets using different segmentation backbones. Seg.Back.: Network backbone for segmentation. Sup.: supervision. I: image-level labels. S: Saliency maps. $^*$: Our reimplemented results using official code.}
% The best and the second best are in bold and underlined, respectively.
%Top-1 and Top-2 results are shown in bold and underlined, respectively.
% The top three results are in \textcolor{red}{red}, \textcolor{green}{green} and \textcolor{blue}{blue}, respectively.
\label{tab:sota_res38}
\vspace{-1mm}
\end{table}

%% file: tab/sota_coco.tex
\begin{table}[t]
\small
\centering
\resizebox{1.0\linewidth}{!}{
\renewcommand{\arraystretch}{1.2}
\begin{tabular*}{8.9cm}{@{\extracolsep{\fill}}lccc}
\toprule%[1pt] 
 Method            & Seg.Back.  &Sup.          & Val                       \\ \midrule
% SEC (CVPR16) \cite{kolesnikov2016seed} & VGG16 & I+S &22.4 \\
% DSRG (CVPR18) \cite{huang2018weakly} & VGG16 & I+S&26.0 \\
EPS (CVPR21) \cite{WSSS_railroad_2021_CVPR} & ResNet101 & I+S & 35.7 \\
AuxSegNet (ICCV21) \cite{WSSS_LAT_2021_ICCV} & ResNet38 & I+S & 33.9 \\
ESOL (NeurIPS22)~\cite{WSSS_expand_sh_2022_NIPS} & ResNet101 &I+S &42.6 \\
L2G (CVPR22) \cite{WSSS_L2G_2022_CVPR} & ResNet101 & \textbf{I+S} & \textbf{44.2} \\ \midrule
Wang \etal (IJCV20) \cite{WSSS_IAL_2020_IJCV}&VGG16&I&27.7\\
SEAM (CVPR20) \cite{WSSS_SEAM_2020_CVPR} & ResNet38 & I&31.9 \\
CONTA (NeurIPS20) \cite{WSSS_causal_2020_NIPS} & ResNet38 & I&32.8 \\
CDA (ICCV21) \cite{WSSS_CDA_2021_ICCV} & ResNet38 & I &33.2 \\
SIPE (CVPR22)  \cite{WSSS_SIPE_2022_CVPR}& ResNet101 & I &40.6\\
MCTformer (CVPR22)  \cite{WSSS_MCT_2022_CVPR}& ResNet38 & I &42.0\\
 \midrule
 \textbf{\nn} (Ours) & ResNet38 &I& 42.7 \\
\textbf{\nn} (Ours) & ResNet101 &\textbf{I}& \textbf{44.3} \\
  \bottomrule%[1pt]  
\end{tabular*}
}
\vskip 0.1in
\caption{Performance comparison of WSSS methods in terms of mIoU (\%) on the MS COCO~\cite{dataset_coco_2014_ECCV} \textit{val} set. }
%\vspace{-5mm}
\label{tab:coco}
%\vspace{-10pt}
\label{tab:segsota_coco}
% \end{center}
% \vspace{-4ex}
\end{table}

%% file: tab/smooth_sharpen.tex
\begin{table}[t]
 %\footnotesize
 \small
 \newcommand{\CC}[1]{\cellcolor{gray!#1}}
%\vspace{-0.38cm}
\centering
		\setlength{\tabcolsep}{2mm}{
			\scalebox{0.95}{
				\begin{tabular}{c|cccccc}
					\toprule
					Temperature $\tau_{1}$ & 0.5    & 1 & 2 & 25 & 50 & 75    \\
					\midrule
                   \multicolumn{1}{l}{Trans.USAGE} & \multicolumn{3}{c} {\emph{sharpening}} &\multicolumn{3}{c}{}
                    \\
					\midrule
					mIoU (\%)    & \CC{15}67.2 & \CC{15}\textbf{67.7}   & \CC{15}67.1   & 52.3     & 51.4 & 50.3
                    \\ \midrule %\bottomrule					\midrule
                   \multicolumn{1}{l}{CNN.USAGE} & \multicolumn{3}{c}{} &
                   \multicolumn{3}{c}{\emph{smoothing}}
                    \\
					\midrule
					mIoU (\%)   & 48.9 & 49.6   & 50.5    & \CC{15}57.5     & \CC{15}\textbf{57.7} & \CC{15}56.7\\ \bottomrule
	\end{tabular}}}
 \vskip 0.1in
		\caption{Discussion of the temperature parameter $\tau_{1}$ following the deterministic principle, \emph{i.e.,} smoothing for CNNs or sharpening for Transformers.}
  \label{tab_smooth_sharpen}
	\vspace{-0.3cm}
	%\vspace{-0.4cm}
\end{table}

%% file: tab/abla_real4.tex
\begin{table}[t]
\newcommand{\CC}[1]{\cellcolor{gray!#1}}
\renewcommand{\arraystretch}{0.01}

 %\scriptsize
 \small
%\vspace{-0.33cm}
\centering
   \tabcolsep=0.18cm
				\begin{tabular}{c|ccccccc}
			\toprule
			%Regularzation scaling & Student branch & Teacher branch & mIoU (train)    \\ \midrule
             $\gamma_t$ &0.05 &0.05&0.05 &0.05&0.05&0.05 &0.05\\ \midrule
             
             $\gamma_s$ &0.05 &0.10&0.15 &0.2&0.25&0.15 &0.15 \\ \midrule
             $\delta_s$ &0.01 &0.01&0.01 &0.01&0.01&0  &0.02 \\ \midrule
             mIoU~(\%) &66.4 &67.2 &\CC{15}\textbf{67.7} &67.3&66.9&67.0 &66.8    \\ \bottomrule
	\end{tabular}	
  \vskip 0.1in
 \caption{Ablation on different rates of drop path and dropout for our network adjustment, on the PASCAL VOC~\cite{dataset_pascal_2010_IJCV} \emph{train} set. $\gamma_t$: drop path rate in the teacher model (default setting). $\gamma_s$: drop path rate in the student model. $\delta_s$: drop rate in the student model.}
 \label{abla_rate4}
	%\vspace{-0.2cm}
	%\vspace{-0.47cm}
\end{table}

%% file: tab/abla_sa.tex
% \begin{table} 
% \newcommand{\CC}[1]{\cellcolor{gray!#1}}
% \small
% \begin{center}
%        %\begin{adjustbox}{width=0.95\linewidth}
% 	\begin{tabular}{c|ccc}

% 		\toprule
%     Function & Constant & Linear& Self-adaptive \\ \midrule
% \multicolumn{1}{l}{} &\multicolumn{3}{c}{\emph{Trans.USAGE}} \\ \midrule 
%     mIoU(\%) &    66.6 & 67.2 & \CC{15}\textbf{67.7}    \\ \midrule     
% \multicolumn{1}{l}{} &\multicolumn{3}{c}{\emph{CNN.USAGE}}  \\ \midrule 
%     mIoU(\%) &    55.5 & 56.9 & \CC{15}\textbf{57.7}   \\      

% % &  0.0 &    0.99 & 0.5 & 86.61 & 0.62\\ 

% 		              \bottomrule  
%                 \end{tabular}
%                 \end{center}

%  \caption{Discussion of the self-adaptive strategy for our network adjustment, on the PASCAL VOC~\cite{dataset_pascal_2010_IJCV} \emph{train} set. Here, we present growth trends for three functions that determine changes in network adjustment for the student. The term ``Constant'' refers to a fixed value without any increase, ``Linear'' represents a linear function, and ``Self-adaptive'' indicates the self-adaptive strategy.}
%   \label{ab_sa}
%     %\vspace{-20pt}
    
% \end{table}  

\begin{table} 
\newcommand{\CC}[1]{\cellcolor{gray!#1}}
\small
\begin{center}
       %\begin{adjustbox}{width=0.95\linewidth}
	\begin{tabular}{c|ccc}

		\toprule
    Strategy & Fixed & Linear& Self-adaptive \\ \midrule
\multicolumn{1}{l}{} &\multicolumn{3}{c}{\emph{Trans.USAGE}} \\ \midrule 
    mIoU(\%) &    66.6 & 67.2 \color{Highlight}($+$0.6) & \CC{15}\textbf{67.7} \textbf{\color{Highlight}($+$1.1)}   \\ \midrule     
\multicolumn{1}{l}{} &\multicolumn{3}{c}{\emph{CNN.USAGE}}  \\ \midrule 
    mIoU(\%) &    55.5 & 56.9 \color{Highlight}($+$1.4) & \CC{15}\textbf{57.7} \textbf{\color{Highlight}($+$2.2)}  \\      

% &  0.0 &    0.99 & 0.5 & 86.61 & 0.62\\ 

		              \bottomrule  
                \end{tabular}
                \end{center}

 \caption{Discussion of the introduced self-adaptive strategy for our network adjustment, on the PASCAL VOC~\cite{dataset_pascal_2010_IJCV} \emph{train} set. Here, we present three strategies that determine the change of the adjustment gap between weak and strong network adjustments. The term ``Fixed'' refers to a fixed value, ``Linear'' represents a linear function, and ``Self-adaptive'' indicates the self-adaptive strategy.}
  \label{ab_sa}
    \vspace{-3pt}
    
\end{table}

%% file: tab/abla_variants.tex
\begin{table}[t]
\newcommand{\CC}[1]{\cellcolor{gray!#1}}
\small
\centering
\resizebox{1.0\linewidth}{!}{
\setlength{\tabcolsep}{0.12cm}
\begin{tabular*}{8.6cm}{@{\extracolsep{\fill}}cccc|cccc|c}
%\begin{tabular}{lc}
\toprule[1pt]

\multicolumn{4}{c|}{Mapping} &\multicolumn{4}{c|}{Regularization} & mIoU (\%) \\ \midrule
CAM &  MCT  & MIL  & CM & w/o & GT & OAA &  AA & \\ \midrule[1pt]
%TS-CAM~\cite{gao2021ts} &29.9\\
%TS-CAM$^\ast$~\cite{gao2021ts} & 41.3 \\
%\nn-V1 (Attention) & 47.2 \\
%\nn-V1 (Attention + PatchAffinity) & 55.2 \\
\multicolumn{9}{c}{\emph{Seed area generation from CNN}} \\
\cmark   &  &  & & \cmark & & & & 48.7 \\ 
   &  & \cmark & & \cmark & & & & 29.3 \\ 
\cmark   &  &  & & &  & \cmark & & 53.9 \\ 
\cmark   &  &  & & & \cmark & & & 55.4 \\ 
%\rowcolor{gray}
   &  &  & \cmark & & & & \cmark & \CC{15}\textbf{57.7} \\ \midrule
\multicolumn{9}{c}{\emph{Seed area generation from Transformer}} \\
\cmark   &  &  & & \cmark & &&  & 52.3\\ 
  & \cmark  &  & &  & & & &  55.2 \\ 
&  & \cmark & & \cmark & & & & 54.3 \\ 
&  &  & \cmark & & \cmark & & & 62.4 \\
\cmark   &  &  & & &  & \cmark & & 65.1 \\ 
\cmark &  &  &  & & & & \cmark & 56.5 \\ 
%\rowcolor{gray}
&  & & \cmark & & & & \cmark & \CC{15}\textbf{67.7}\\
%29.4& 54.3& 55.2 &55.4 &62.4 & \CC{15}\textbf{66.6} &  \\
%\cmark& &  &  & & 57.6 \\
%& \cmark &  &  & & 29.4 \\
%& & \cmark &   & &58.3 \\
% & &  & \cmark &  & 61.7\\
% &  &  & & &\cmark \CC{15}\textbf{66.6} \\
%\nn &61.1\\
%\nn + PatchAffinity & \textbf{66.3} \\ 
\bottomrule[1pt]
\end{tabular*}
}
\vskip 0.1in
\caption{Comparison of the variants of the USAGE on the PASCAL VOC 2012~\cite{dataset_pascal_2010_IJCV} \textit{train} set.
MCT: Class-to-patch attention map~\cite{WSSS_MCT_2022_CVPR}. MIL: MIL-based method~\cite{WSSS_FIP_2015_CVPR}. GT: cross-view regularization by geometric transform~\cite{WSSS_SEAM_2020_CVPR}. CM:  Activation shape controlling based mapping function in our USAGE. OAA: cross-view regularization by online attention accumulation~\cite{WSSS_OAA_2019_ICCV}. AA: Activation shape regularization by architecture adjustment in our USAGE. Mapping: The mapping function optimized by the first term in Eq.~\ref{eq:USAGE_optimization}. Regularization: The second term in Eq.~\ref{eq:USAGE_optimization}.}%Controlling: Activation shape controlling. Regularization: Activation shape regularization. }
%``MCT'' denotes the class-to-patch attention map proposed in~\cite{WSSS_MCT_2022_CVPR}. ``MIL'' denotes the MIL-based method~\cite{WSSS_FIP_2015_CVPR}. ``IT'' denotes the cross-view regularization by image transform~\cite{WSSS_SEAM_2020_CVPR}. }%$^\dag$ means apply the MIL-based method~\cite{WSSS_FIP_2015_CVPR} on the Transformer. }
%\vspace{-4pt}
\label{tab:abla_variants}
% \end{center}
%\vspace{-4mm}

\end{table}

%% file: sec/5_conclusions.tex
\section{Conclusion}

We developed a unified optimization paradigm for generating seed areas, called USAGE. The objective function of USAGE consists of two terms: a generation loss that enables seed areas to be adaptive to different types of networks via a simple temperature parameter, and a regularization loss ensures the consistency between the seed areas that are generated by self-adaptive network adjustment from different views. By incorporating these two terms, USAGE can solve problematic activations of seed areas for both CNNs and Transformers. Experimental results demonstrated that our \nn~could bring continuous performance gains on seed area generation and achieved state-of-the-art results on PASCAL VOC and COCO datasets. 

\noindent\textbf{Limitation.} Throughout this paper, we investigated the step-wise pipeline of WSSS, while the end-to-end pipeline that often produces inferior results remains uncovered. In the future, we look forward to generalizing the idea of USAGE to improve the end-to-end WSSS algorithms.

\noindent\textbf{Acknowledge} 
This work was supported by NSFC 62176159, Natural Science Foundation of Shanghai 21ZR1432200, Shanghai Municipal Science and Technology Major Project 2021SHZDZX0102, and the Fundamental Research Funds for the Central Universities.